\documentclass[oneside]{amsart}

\usepackage[british]{babel}
\usepackage{listings,bm,tikz,amsmath,xcolor,bm,tikz,amssymb,pgfplots,fourier,algorithm,algorithmic,amsmath,blkarray,xparse,soul,comment,url,hyperref,subfigure,fancyhdr, amsaddr}

\usepackage[a4paper, total={6in, 8.5in}]{geometry}

\fancypagestyle{mystyle}{
  \fancyhf{} 

  \fancyhead[L]{\scriptsize\nouppercase{\rightmark}}
  \fancyhead[R]{\thepage}

}

\usepgfplotslibrary{statistics}
\usetikzlibrary{arrows,automata,circuits,plotmarks,decorations.markings,trees,shapes,pgfplots.groupplots}
\pgfplotsset{compat=1.17}
\tikzset{dot/.style = {circle, fill, minimum size=#1,inner sep=0pt, outer sep=0pt, fill, circle},dot/.default = 6pt}
\tikzstyle{a}=[->,>=stealth,dashed]
\tikzstyle{a2}=[->,>=stealth]
\tikzstyle{a3}=[<->,>=stealth]
\pgfplotsset{asse/.style={xlabel=$\mathrm{FRL}$,xmin=0.0,xmax=1,ymin=0,ymax=1,xtick={0,0.25,0.5,0.75,1},ytick={0,0.25,0.5,0.75,1},grid=both}}
\pgfplotsset{brier/.style={black,thick,densely dotted}}
\pgfplotsset{acc2/.style={fill=black!20,draw=black,opacity=0.7,const plot,ybar interval}}
\pgfplotsset{acc1/.style={fill=black!60,draw=black,opacity=0.7,const plot,ybar interval}}

\newtheorem{proposition}{Proposition}
\newtheorem{proofprop}{Proof of Proposition}

\theoremstyle{definition}

\theoremstyle{remark}

\numberwithin{equation}{section}

\makeatletter
\renewcommand{\email}[2][]{%
  \ifx\emails\@empty\relax\else{\g@addto@macro\emails{,\space}}\fi%
  \@ifnotempty{#1}{\g@addto@macro\emails{\textrm{(#1)}\space}}%
  \g@addto@macro\emails{#2}%
}
\makeatother

\begin{document}
\title{On the Correlation between Individual Fairness and Predictive Accuracy in Probabilistic Models}

\author{Alessandro Antonucci$^*$}
\address[1]{IDSIA USI-SUPSI, Lugano, Switzerland}
\curraddr{}
\email{alessandro.antonucci@idsia.ch}
\thanks{$^\ast$Corresponding author.}

\author{Eric Rossetto}
\address{IDSIA USI-SUPSI, Lugano, Switzerland}
\curraddr{}
\email{eric.rossetto@idsia.ch}
\thanks{}

\author{Ivan Duvnjak}
\address{SUPSI, Lugano, Switzerland}
\curraddr{}
\email{ivan.duvnjak@supsi.ch}
\thanks{}


\date{16/09/2025}

\dedicatory{}

\begin{abstract}
We investigate individual fairness in generative probabilistic classifiers by analysing the robustness of posterior inferences to perturbations in private features. Building on established results in robustness analysis, we hypothesise a correlation be\-tween robustness and predictive accuracy—specifically, instances exhibiting greater robustness are more likely to be classified accurately. We empirically assess this hypothesis using a benchmark of fourteen datasets with fairness concerns, employing Bayesian networks as the underlying generative models. To address the computational complexity associated with robustness analysis over multiple private features with Bayesian networks, we reformulate the problem as a \emph{most probable explanation} task in an auxiliary Markov random field. Our experiments confirm the hypothesis about the correlation, suggesting novel directions to mitigate the traditional trade-off between fairness and accuracy.
\end{abstract}

\maketitle

\section{Introduction}\label{sec:introduction}
The widespread adoption of machine learning and knowledge-based systems for decision support and recommendation has made the development of tools and metrics for algorithmic fairness increasingly urgent \cite{mitchell2021algorithmic}. Algorithmic fairness generally refers to the identification and mitigation of biases in model predictions with respect to sensitive, or \emph{private}, features. However, simply omitting these private features from the model is often insufficient: it can significantly degrade predictive accuracy and fails to guarantee fairness, as other features may remain highly correlated with the excluded ones \cite{cummings2019compatibility}.

In this paper, we focus on generative probabilistic models that produce predictions in the form of posterior distributions over the target variable, conditioned on a complete assignment of both private and public features. For such models, biases can be naturally characterized through the robustness of predictions to perturbations in private features. We demonstrate how this notion of robustness aligns with established formulations of inference under non-ignorable missingness \cite{decooman2004updating}. Prior work has further shown that robustness can be positively correlated with predictive accuracy \cite{llerena2021}. This connection is particularly compelling, as it suggests a mitigation strategy for the classical trade-off between fairness and accuracy \cite{cooper2021emergent}: at the \emph{individual} level \cite{dwork2012fairness}, fair classifications may also be accurate, allowing standard predictive models to be employed without modification for the robust instances.

To validate such a conjecture we consider an extensive benchmark including fourteen datasets related to fairness. In our study, we adopt Bayesian networks as probabilistic generative models, compute the robustness, based on a fairness deviation measure \cite{woodworth17a} for each test instance, and evaluate the correlation with the predictive performance by using the Brier score. The empirical results we observe are in line with those reported in the earlier literature on robustness (e.g., \cite{mattei2020}):  the most robust, and hence, in our setup, fair, instances are also ``easier'' to classify. To the best of our knowledge, our work is the first attempt to apply concepts from robust probabilistic methods to fairness analysis. 

We leverage this novel connection to offer a more interpretable perspective on fairness. Future work will investigate how this link might be harnessed to actively control fairness during training, as well as evaluate its effectiveness in comparison to existing methods (e.g., \cite{xu2021}).

The robustness measure we adopt could, in principle, require a brute-force computation over all possible joint assignments of the private features. However, we reformulate the inference as a \emph{most probable explanation} (MPE) task within an auxiliary Markov random field derived from the original Bayesian network. This reformulation yields computational benefits, which we both formally establish and validate empirically.

The code used for the experiments is available for reproducibility in a dedicated Github repository.\footnote{\href{https://github.com/IDSIA-papers/fairness}{github.com/IDSIA-papers/fairness}.\label{fn:github}} The repository includes references to the original datasets, preprocessing scripts, discretised versions of the datasets, and a custom implementation of some inference algorithms. 

The paper is organised as follows. In Section~\ref{sec:basics}, we define the necessary background on probabilistic graphical models. Our approach to fairness analysis is discussed in Section~\ref{sec:fairness}. The mapping to Markov random fields is derived in Section~\ref{sec:mrf}. The benchmark we adopt is detailed in Section~\ref{sec:benchmark}, while the results of our experiments are in Section~\ref{sec:experiments}. A critical discussion on the limitations of the present work and possible strategies to bypass them is in Section ~\ref{sec:limitations}, while our conclusions are in Section~\ref{sec:conclusions}. Technical results and additional experimental analyses are gathered in the appendix.

\section{Basics}\label{sec:basics}
Let us first review the necessary background information on probabilistic graphical models. We remind the reader to \cite{koller2009} for deeper insights. 

\paragraph{Variables, Potentials, and Tables.}
We denote variables by uppercase letters and their states with lowercase letters. A subscripted variable in $\Omega$ is used instead for the sample space. Thus $v \in \Omega_{V}$ is a generic state of the variable $V$. We focus on discrete variables only. A non-negative function $\phi:\Omega_{V}\to\mathbb{R}$ is called a \emph{potential} over $V$. A potential over $V$ that is also normalised to one is called \emph{probability mass function} (PMF) and denoted as $P(V)$. PMF $P$ gives the probability $P(v)$ to $V=v$ for each $v\in\Omega_{V}$. A \emph{conditional probability table} (CPT) for $V$ given $W$ is a collection of PMFs over $V$ indexed by the values of $W$ and it is denoted as $P(V|W)$. We also define potentials over joint variables. As an example, the CPT $P(V|W)$ is a special example of a multi-variate potential $\phi(V,W)$.

\paragraph{Bayesian Networks (BNs).} Consider a set of variables $\bm{V}$. A BN $B$ over $\bm{V}$ is a collection of CPTs, one for each $V\in\bm{V}$, say  $\{P(V|\mathrm{Pa}_V)\}_{V\in\bm{V}}$ where $\mathrm{Pa}_V \subseteq \bm{V} \setminus \{V\}$ for each $V\in\bm{V}$. From the BN specification $B$, we induce a directed graph $\mathcal{G}_B$ whose nodes are in one-to-one correspondence with the variables in $\bm{V}$ and such that there are directed arcs from $W$ to $V$ for each $W\in\mathrm{Pa}_V$ and $V\in\bm{V}$. Here we only consider \emph{acyclic} BNs, i.e., BNs such that $\mathcal{G}_B$ does not contain directed loops.  
An acyclic BN $B$ compactly defines a joint PMF $P(\bm{V})$ as follows:
\begin{equation}\label{eq:joint}
    P(\bm{v})=\prod_{V\in\bm{V}} P(v|\mathrm{pa}_V)\,,
\end{equation}
for each $\bm{v}\in\Omega_{\bm{V}}$, where the values $v$ and $\mathrm{pa}_V$ are those consistent with $\bm{v}$. 
Given a dataset of i.i.d. joint observations of $\bm{V}$, there are various \emph{structural learning} techniques to obtain a directed acyclic graph over the variables. Typical approaches are maximising likelihood-based scores. Once the graph is detected, the BN CPTs can be quantified from the data by means of a count function $n$, i.e.,
\begin{equation}\label{eq:laplace}
P(v|\mathrm{pa}_V) = \frac{ n(v,\mathrm{pa}_V)+\frac{s}{|\Omega_{V}|}}{\sum_{v\in\Omega_{V}}n(v,\mathrm{pa}_V)+s}\,,
\end{equation}
where a Laplace smoothing with equivalent sample size $s>0$ is preferred to a frequentist estimate.

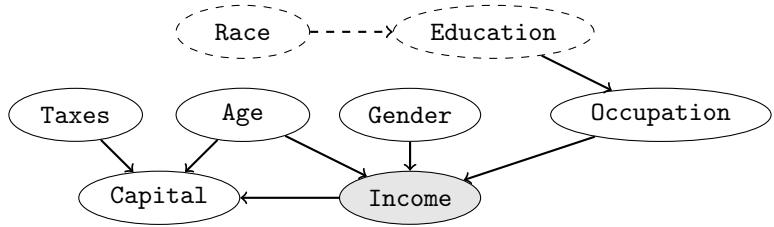
\begin{figure}[htp!]
\centering
\tikzstyle{every node}=[draw, ellipse, minimum height=20pt,minimum width=50pt,align=center]
\tikzstyle{target}=[fill=black!10]
\tikzstyle{out2}=[dashed]
\begin{tikzpicture}[scale=1.1]
\node[target] (I) at (0,0) {\tt Income};
\node (A) at (-2,1) {\tt Age};
\node (T) at (-4,1) {\tt Taxes};
\node[out2] (R) at (-2,2) {\tt Race};
\node (G) at (0,1) {\tt Gender};
\node (C) at (-3,0) {\tt Capital};
\node[out2] (E) at (1,2) {\tt Education};
\node (O) at (3,1) {\tt Occupation};
\draw[->,thick]{ (I) -> (C)};
\draw[->,thick]{ (A) -> (C)};
\draw[->,thick]{ (A) -> (I)};
\draw[->,thick]{ (O) -> (I)};
\draw[->,thick,dashed]{ (R) -> (E)};
\draw[->,thick]{ (E) -> (O)};
\draw[->,thick]{ (G) -> (I)};
\draw[->,thick]{ (T) -> (C)};
\end{tikzpicture}
\caption{The directed graph of a BN. The nodes outside the Markov blanket of node {\tt Income} are dashed.}
\label{fig:bn}
\end{figure}

Given a node $W$, the \emph{Markov blanket} $\bm{V}_W$ of $W$ is a subset of $\bm{V}$ including the variable $W$, its parents $\mathrm{Pa}_W$, the children $\mathrm{Ch}_W$ of $W$, and the parents of those children. We denote as $\mathcal{G}_W$ the subgraph of $\mathcal{G}$ including only the variables in $\bm{V}_W$. An example is in Figure~\ref{fig:bn}. A BN over $\bm{V}_W$ and based on $\mathcal{G}_W$ can be obtained by taking the CPTs of $W$ and its children from $B$ and arbitrarily setting the marginal PMFs for the parents of $W$ and the parents of its children. In the following we will use uniform PMFs for these marginals and denote such a BN as $B_W$.

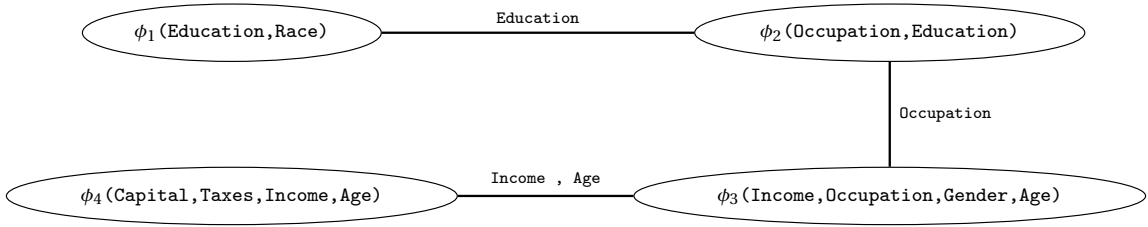
\begin{figure}[htp!]
\centering
\resizebox{\textwidth}{!}{
    \tikzstyle{aa}=[draw,ellipse,minimum height=20pt,align=center]
    \begin{tikzpicture}
    \node[aa] (v1) at (-4,2) {\tt\scriptsize$\phi_1$(Education,Race)};
    \node[aa] (v2) at (4,2) {\tt\scriptsize$\phi_2$(Occupation,Education)};
    \node[aa] (v3) at (4,0) {\tt\scriptsize$\phi_3$(Income,Occupation,Gender,Age)};
    \node[aa] (v4) at (-4,0) {\tt\scriptsize$\phi_4$(Capital,Taxes,Income,Age)};
    \draw[-,thick]{(v1) edge node[above]{\tt\tiny Education} (v2)};
    \draw[-,thick]{(v2) edge node[right]{\tt\tiny Occupation} (v3)};
    \draw[-,thick]{(v3) edge node[above]{\tt\tiny Income , Age}  (v4)};
    \end{tikzpicture}
}
\caption{The undirected graph of a MRF. Labels over edges highlight the variables shared among potentials.}
\label{fig:mrf}
\end{figure}

\paragraph{Markov Random Fields (MRFs).}
A MRF $M$ over $\bm{V}$ is a collection of $k$ potentials $\{\phi_i(\bm{V}_i)\}_{i=1}^k$, with $\bm{V}_i \subseteq \bm{V}$ for each $i=1,\ldots,k$ and such that for each $V\in\bm{V}$ there is at least an index $i$ for which $V\in\bm{V}_i$. Given the MRF $M$, we induce an undirected graph $\mathcal{G}_M$ whose $k$ nodes are in correspondence with the joint variables $\{\bm{V}_i\}_{i=1}^k$ and such that there is an edge between two nodes if and only if the corresponding joint variables share at least one variable. An example is in Figure~\ref{fig:mrf}. A MRF defines a joint PMF $P(\bm{V})$ as follows:
\begin{equation}\label{eq:mrf}
P(\bm{v}) \propto \prod_{i=1}^n \phi_i(\bm{v}_i)\,,
\end{equation}
for each possible value $\bm{v}$ of $\bm{V}$, where the values of $\bm{v}_i$ are those consistent with $\bm{v}$.

A BN can be intended as an MRF whose potentials are the CPTs. Because of Eq.~\eqref{eq:joint}, for those models, there is no normalisation constant to compute in Eq.~\eqref{eq:mrf}. As an example the MRF induced by the BN in Figure~\ref{fig:bn} is based on the undirected graph in Figure~\ref{fig:mrf}.

\paragraph{Inference.} A typical inference tasks in MRFs and BNs is \emph{updating}. This consists in the computation of the posterior PMF $P(V_Q|\bm{v}_E)$ for a queried variable $V_Q\in\bm{V}$, given an evidence about some other variables $\bm{V}_E=\bm{v}_E$ with $\bm{V}_E\subseteq \bm{V}\setminus\{V_Q\}$. \emph{Marginal MAP} (MMAP) consists instead of finding the most probable configuration of a joint queried variable $\bm{V}_Q \subseteq \bm{V}$ given the evidence, i.e.,
\begin{equation}\label{eq:mpe}
\bm{v}_Q^* = \arg \max_{\bm{v}_Q} P(\bm{v}_Q|\bm{v}_E)\,.
\end{equation}
If, in particular, $\bm{V}=(\bm{V}_Q,\bm{V}_E)$, i.e., all the variables are either queried either observed, the task is called \emph{most probable explanation} (MPE). Moreover, notice that the maximisation in Eq.~\eqref{eq:mpe} is with respect to $\bm{v}_Q$, the probability of the evidence $P(\bm{v}_E)$ is a positive constant and the task in Eq.~\eqref{eq:mpe} can be equivalently solved by replacing the conditional probability in the r.h.s. with a joint probability. 

Both updating and MPE tasks can be solved exactly by \emph{variable elimination} schemes. Given an elimination order over $\bm{V}$, this consists in processing the potential obtained by multiplying all the potentials, including the variable we want to eliminate. The processing consists in the marginalisation of the variables not involved in the query, the instantiation of the observed variables and the queries of the updating task, and the maximisation of the queried variables of the MPE. The maximum \emph{arity} of the potentials created in this way is called \emph{treewidth}. Inference with this scheme is exponential with respect to this parameter.

\section{(Probabilistic) Fairness Analysis}\label{sec:fairness} 
We are interested in evaluating the presence of biases in a training dataset $\mathcal{D}$ over the discrete variables $\bm{V}:=(Y,\bm{X},\bm{Z})$. The dataset refers to a classification task where the target variable is $Y$ and the features are distinguished between private, those in $\bm{X}$, and public, those in $\bm{Z}$. A test dataset $\mathcal{D}'$ over the same variables is also assumed available. We learn from $\mathcal{D}$ a generative probabilistic model in the form of a joint PMF $P(Y,\bm{X},\bm{Z})$. In our classification task, we want to identify the most probable value of $Y$ given a test instance $(\hat{\bm{x}},\hat{\bm{z}})$ of the features taken from $\mathcal{D}'$, i.e.,
\begin{equation}\label{eq:classification}
    y^* := \arg \max_{y\in\Omega_Y} P(y|\hat{\bm{x}},\hat{\bm{z}})\,,
\end{equation}
where a zero-one loss function is considered just for the sake of light notation. Besides achieving a good predictive level on the test instances, we might be interested in bounding the sensitivity of inferences to changes in the values of the private features and, ideally, keep these changes as small as possible. This could be done at the predictive level, i.e., checking whether the same class as in Eq.~\eqref{eq:classification} is obtained for perturbed values of the private features $\hat{\bm{x}}$. However, a deeper analysis can be achieved by a dissimilarity measure $\delta$ for the PMFs over the target $Y$. We thus consider the following optimisation:
\begin{equation}\label{eq:delta}
\bm{x}^*:=\arg \max_{\bm{x}\in\Omega_{\bm{X}}} \delta\left[P(Y|{\bm{x}},\hat{\bm{z}}),P(Y|\hat{\bm{x}},\hat{\bm{z}}) \right]\,,
\end{equation}
and call \emph{fairness robustness level} (FRL) of the instance $(\hat{\bm{x}},\hat{\bm{z}})$ the corresponding maximum, i.e.,
\begin{equation}\label{eq:frl}
\rho(\hat{\bm{x}},\hat{\bm{z}}):=\delta \left[ P(Y|\bm{x}^*,\hat{\bm{z}}) , P(Y|\hat{\bm{x}},\hat{\bm{z}}) \right]\,.
\end{equation}
It is a simple remark that $\rho=0$ implies a context-specific irrelevance of the private features to the target given a particular instance of the public features.

Let us consider, for the sake of simplicity, the case of a Boolean target $Y$ and take as distance $\delta$ the Manhattan distance, i.e., 
\begin{equation}\label{eq:manhattan}
\delta \left[ P(Y),P'(Y) \right]:=\frac{|P(y_0)-P'(y_0)|+|P(y_1)-P'(y_1)|}{2}=|P(y_0)-P'(y_0)|\,,
\end{equation}
where $y_0$ is a short form for $Y=0$ and similarly for $y_1$. Because of Eq.~\eqref{eq:manhattan}, Eq.~\eqref{eq:delta} rewrites as:
\begin{equation}\label{eq:max}
\bm{x}^*:=\arg \max_{\bm{x}\in\Omega_{\bm{X}}} \left| P(y_0|\bm{x},\hat{\bm{z}})-P(y_0|\hat{\bm{x}},\hat{\bm{z}}) \right|\,.
\end{equation}
It is a trivial exercise to notice that considering $y_1$ instead of $y_0$ in Eq.~\eqref{eq:max} would not affect $\bm{x}^*$. This implies that the FRL corresponds to the \emph{fairness deviation measure} often used in the literature to rank unfairness \cite{konstantinov2021fairness}. Rather than directly addressing the optimisation in Eq.~\eqref{eq:max}, we can focus on the two following optimisations:
\begin{eqnarray}
 \overline{\bm{x}} &=& \arg \max_{\bm{x}\in\Omega_{\bm{X}}} P(y_0|\bm{x},\hat{\bm{z}})\,, \label{eq:cir1}\\
\underline{\bm{x}} &=& \arg \min_{\bm{x}\in\Omega_{\bm{X}}} P(y_0|\bm{x},\hat{\bm{z}})\,. \label{eq:cir2}   \end{eqnarray}
This is possible because of the following result whose proof is in the appendix.
\begin{proposition}\label{prop:x}
The solutions of the optimisations in Eqs.~\eqref{eq:max}, \eqref{eq:cir1} and \eqref{eq:cir2} are linked by the following relation:
\begin{equation}\label{eq:xstar}
    \bm{x}^*:= \left\{ \begin{array}{ll} \overline{\bm{x}}  & \mathrm{if} \, P(y_0|\hat{\bm{x}},\hat{\bm{z}}) < \frac{1}{2}\left[{P(y_0|\overline{\bm{x}},\hat{\bm{z}})+
    P(y_0|\underline{\bm{x}},\hat{\bm{z}})} \right] \,,\\ \underline{\bm{x}} & \mathrm{otherwise}\,.\end{array}\right.
\end{equation}
\end{proposition}
It is important to notice that the task in Eq.~\eqref{eq:cir1} is substantially different from an MPE task as in Eq.~\eqref{eq:mpe} because the queried variables in Eq.~\eqref{eq:cir1} are after the conditioning bar (see discussion in Section~\ref{sec:basics}). Thus, while in MPE we can equivalently address the task in the joint model instead of the conditional one becase the evidence is always the same, the same cannot be done with 
Eq.~\eqref{eq:cir1}, and similarly for Eq.~\eqref{eq:cir2}. These tasks can be regarded as a \emph{conservative} inference based on the law proposed in \cite{decooman2004updating}. Yet, no standard BN algorithm can solve the task, while a brute-force approach requires a number of BN inferences exponential with respect to $|\bm{X}|$. In \cite{antonucci2006a}, the MMAP-like optimisation has been mapped to a \emph{credal network} inference, whose complexity might be still demanding \cite{maua2014probabilistic}. Here we leverage on the fact that we cope with complete observations to map the task to a proper MPE task in an auxiliary MRF. This is detailed in the next section.

\section{Robustness of BN Classifiers as a MRF MPE}\label{sec:mrf}
Thanks to Proposition~\ref{prop:x}, we can solve the optimisation in Eq.~\eqref{eq:max} and hence compute the FRL in Eq.~\eqref{eq:frl} by solving the optimisations in Eqs.~\eqref{eq:cir1} and \eqref{eq:cir2}. In this section we show how these BN tasks can be equivalently addressed in an auxiliary MRF $M$ whose specification is discussed here below.

First, it is a trivial exercise to notice that, for the classification task in Eq.~\eqref{eq:classification}, the BN can be equivalently restricted to the BN over the Markov blanket of $Y$. In the following, we therefore assume that the BN coincides with the Markov blanket of $Y$. 

Let $\bm{X}'$ and $\bm{Z}'$ denote, respectively, the private and public parents of $Y$ and by $\bm{X}''$ and $\bm{Z}''$ the private and public children of $Y$. We define a potential $\phi_Y(\bm{X}',\bm{Z}')$ as:
\begin{equation}\label{eq:phiy}
\phi_Y(\bm{x}',\bm{z}'):=\frac{P(y_1|\bm{x}',\bm{z}')}{P(y_0|\bm{x}',\bm{z}')}
\end{equation}
for each $\bm{x}'\in\Omega_{\bm{X}'}$ and $\bm{z}'\in\Omega_{\bm{Z}'}$. Similarly, we define a potential for each $X\in\bm{X}''$ and $Z\in\bm{Z}''$ as follows:
\begin{eqnarray}
\phi_X(x,\mathrm{pa}_X')&:=&\frac{P(x|\mathrm{pa}_X',y_1)}{P(x|\mathrm{pa}_X',y_0)}\,, \label{eq:phix}\\
\phi_Z(z,\mathrm{pa}_Z')&:=&\frac{P(z|\mathrm{pa}_Z',y_1)}{P(z|\mathrm{pa}_Z',y_0)}\,, \label{eq:phiz}
\end{eqnarray}
for each $x\in\Omega_X$, $\mathrm{pa}_X'\in\Omega_{\mathrm{Pa}_{X}'}$, $z\in\Omega_Z$, and $\mathrm{pa}_Z'\in\Omega_{\mathrm{Pa}_{Z}'}$, where $\mathrm{Pa}_X':=\mathrm{Pa}_X \setminus\{Y\}$ and the same notation is also used for $Z$.
Overall, Eq.~\eqref{eq:phiy} together with Eqs.
\eqref{eq:phix} and
\eqref{eq:phiz} define a MRF $\phi(\bm{X},\bm{Z})$ based on $1+|\bm{X}''|+|\bm{Z}''|$ potentials. The following result, whose proof is in the appendix, holds.
\begin{proposition}
Consider the MRF induced by the potentials in Eq.~\eqref{eq:phiy} and Eqs.
\eqref{eq:phix} and \eqref{eq:phiz}. Let $\phi(\bm{X},\bm{Z})$ denote the joint PMF induced by such a MRF. The BN optimisation in Eq.~\eqref{eq:cir2} is equivalent to a MPE task in this MRF, namely:
\begin{equation}
\underline{\bm{x}} = \arg \max_{\bm{x}\in\Omega_{\bm{X}}} \phi(\bm{x},\hat{\bm{z}})\,. \label{eq:mrfcir2}
\end{equation}
For the task in Eq.~\eqref{eq:cir1} we have instead:
\begin{equation}
\overline{\bm{x}}=\arg \min_{\bm{x}\in\Omega_{\bm{X}}} \phi(\bm{x},\hat{\bm{z}})\,. \label{eq:mrfcir1}    
\end{equation}
\end{proposition}
Solving the task in the MRF has the same worst-case complexity, exponential with respect to $|\bm{X}|$, as trying a brute-force in the BN according to Eqs.~\eqref{eq:cir1} and \eqref{eq:cir2}. Yet, generally speaking the treewidth of the MRF can be smaller than $|\bm{X}|$. In particular, any $X''\in\bm{X}''$ that is not a parent of one of the children of $Y$ can be eliminated by processing $\phi_Y$ alone, this potentially inducing a significant reduction of the computational complexity required to solve the task. This claim will be empirically validated in Section~\ref{sec:experiments}.

\section{A Benchmark of Datasets for Fairness Analysis}\label{sec:benchmark}
The main goal of this paper is to investigate whether the FRL defined in Eq.~\eqref{eq:frl} has a relation with the predictive performance of a probabilistic classifier. For a first empirical analysis, we consider fourteen datasets providing a suitable benchmark for fairness-aware machine learning analyses. These datasets have been also considered in \cite{Le_Quy_2022}, where the choice of their private features is discussed and detailed. Table~\ref{tab:datasets} contains some relevant information about the datasets. Apart from {\tt Communities \& Crime}, all datasets have a binary target variable $Y$. The target of that dataset is instead an integer corresponding to the number of violent crimes in a selected area. A binary $Y$ is obtained by a median-based discretisation (sample median of 374 crimes). Continuous features are similarly processed by quantile-based discretisation leading to quaternary variables. Rows with missing values are finally ignored.

For the experiments, we consider a ten-fold cross validation scheme. The folds are obtained by stratification with respect to the target values. For the sake of reproducibility, the code used to process the original datasets as well as the corresponding discretised (and anonymised) datasets are available in the Github repository of the paper (see Footnote 1). The results of our experiments on these data are discussed in the next section.

\begin{table}[htp!]
\centering
\caption{The datasets considered for the experiments. Besides the name of the dataset and the target variable, we report the number of public and private features, the total number of records, the imbalance level $\Delta$ describing the relative occurrence of the least frequent class, and the number $k$ of folds (out of ten) on which fairness is achieved by \emph{design} (see Section~\ref{sec:experiments}).}\label{tab:datasets}
\resizebox{0.85\textwidth}{!}{
    \begin{tabular}{llrrrrr}
    \hline
    Dataset&$Y$&$|\bm{X}|$&$|\bm{Z}|$&$|\mathcal{D}|$&$\Delta$&$k$\\
    \hline
    {\tt Adult}&Income (50k)&3&9&48'842&.24&0\\
    {\tt Bank marketing}&Deposit Subscription&2&13&45'211&.12&5\\
    {\tt Communities \& Crime}&Violent Crimes&1&20&1'994&.50&10\\
    {\tt COMPAS recid.}&Two-year Recidivism&2&8&7'214&.45&1\\
    {\tt COMPAS violent recid.}&Two-year Violent Recidivism&2&8&4'738&.16&3\\
    {\tt Credit card clients}&Default Payment&3&20&30'000&.0005&10\\
    {\tt Diabetes}&Re-admission&1&16&101'766&.46&10\\
    {\tt Dutch census}&Occupation&1&10&60'420&.48&10\\
    {\tt German credit}&Creditability&2&19&1'000&.30&10\\
    {\tt KDD Census-Income}&Income (50k)&2&21&200'000&.07&0\\
    {\tt Law school}&Pass the bar exam&2&7&20'512&.05&1\\ 
    {\tt OULAD}&Final Result&3&9&31'482&.46&10\\
    {\tt Student-Mathematics}&Final Grade&2&31&395&.33&10\\
    {\tt Student-Portuguese}&Final Grade&2&31&649&.15&10\\
    \hline
    \end{tabular}
}
\end{table}

\section{Experiments}\label{sec:experiments}
For a deeper characterisation of the relation between the FRL in Eq.~\eqref{eq:frl} and the predictive performance of the classifier, together with the zero-one accuracy of the single prediction in Eq.~\eqref{eq:classification}, we also consider the more informative \emph{Brier score}, that is defined here as the square of one minus the probability assigned by the classifier to the actual class, i.e., 
\begin{equation}
\beta(\hat{y},\hat{\bm{x}},\hat{\bm{z}}):=[1-P(\hat{y}|\hat{\bm{x}},\hat{\bm{z}})]^2\,.     
\end{equation}
A zero Brier score indicates perfect confidence in the correct class. It is easy to notice that, with binary classes, the prediction is correct if and only if $\beta(\hat{y},\hat{\bm{x}},\hat{\bm{z}})<0.25$.

We adopt BNs as probabilistic classifiers and the \emph{Pyagrum}\footnote{\href{https://pyagrum.readthedocs.io}{pyagrum.readthedocs.io}.} Python library to learn BNs from the data and compute the queries of interest. In particular, we employ a \emph{tabu} local search with list size equal to 10 and a maximum number of consequent iterative changes equal to 100. 

We focus on \emph{individual} fairness by evaluating complete feature instances. In this context (see Section~\ref{sec:mrf}), features that lie outside the Markov blanket of the target variable 
are irrelevant for prediction. When private features are entirely irrelevant to the target, the classifier achieves fairness \emph{by design}, without any trade-off against the accuracy, and the FRL is trivially zero on all test instances. In our cross-validation scheme, with a same dataset, we might obtain different BN topologies for the different folds. The last column of Table~\ref{tab:datasets} reports the number of folds for which the BN is fair by design. 

We compute the Brier score of each test instance. The low number of private features ($|\bm{X}|\leq 3$) allows to obtain the corresponding FRL by a brute-force approach on the BN according to Eq.~\eqref{eq:delta}. Yet, for a comparison, we obtain the same results by also solving the MPE task in the auxiliary MRF as discussed in Section~\ref{sec:mrf}. In the latter case, we use a custom implementation of the variable elimination algorithm, that also allows to solve the minimisation task in Eq.~\eqref{eq:mrfcir1}. To solve also the latter task, we replace each potential with its opposite.


\begin{figure}[htp!]
\centering
\includegraphics[width=\textwidth]{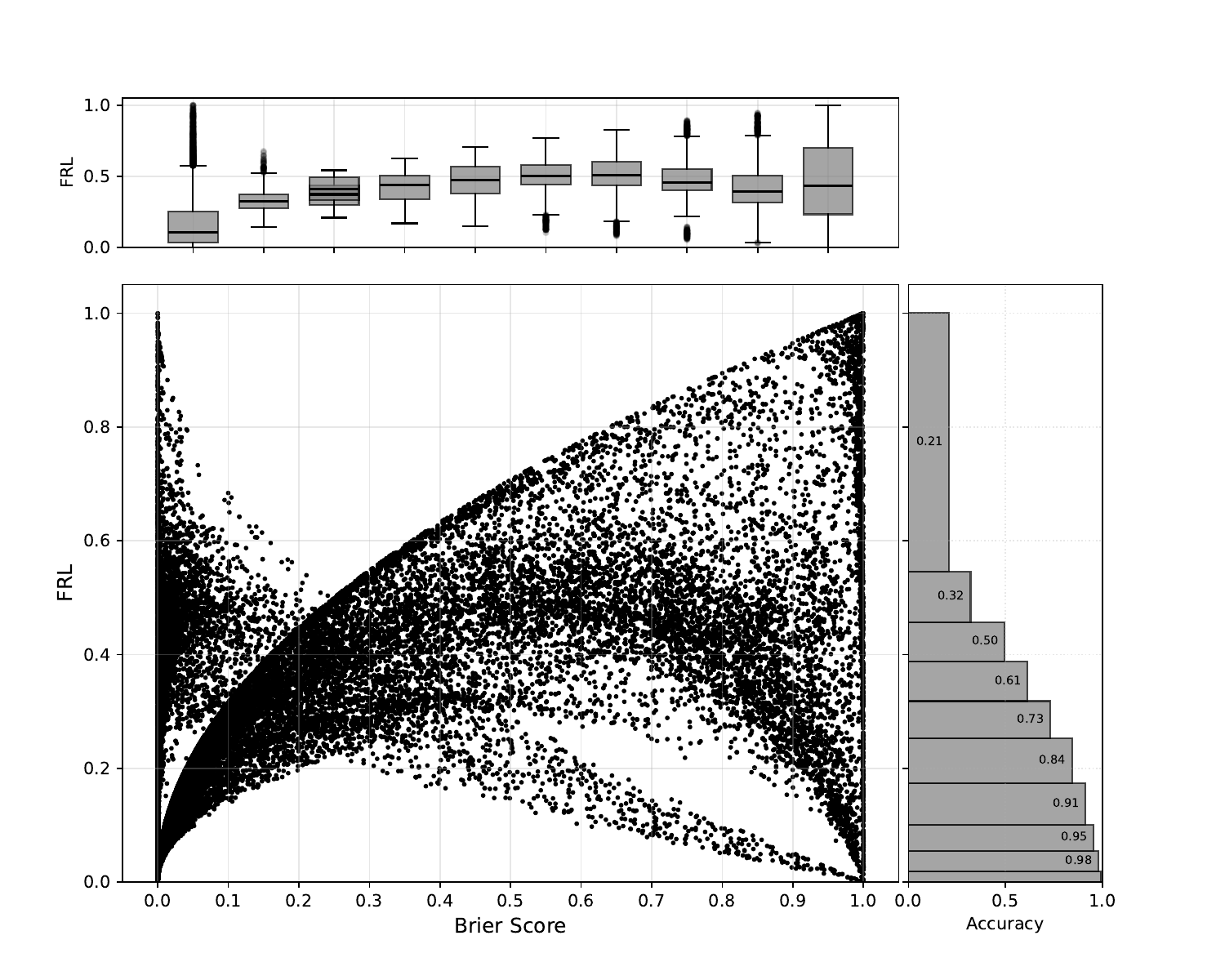}
\caption{Results of the experiments with the {\tt Adult} dataset.\label{fig:adult}}
\end{figure}

A first example of our empirical analysis is provided by Figure~\ref{fig:adult}, where the results for the {\tt Adult} dataset are summarised. The scatter plot displays the Brier score versus the FRL value for each test instance. Instances satisfying $\beta(\hat{y}, \hat{\bm{x}}, \hat{\bm{z}}) < 0.25$—positioned in the left region of the plot—correspond to cases that are accurately classified (approximately $84\%$ of the total sample). For these points, because of Eq.~\eqref{eq:max}, the FRL cannot exceed the value of $P(\hat{y}|\hat{\bm{x}},\hat{\bm{z}})=1-\sqrt{\beta(\hat{y},\hat{\bm{x}},\hat{\bm{z}})}$. This explain why all these points are located under a parabolic curve. Similar considerations can be done for the incorrectly classified instances corresponding to $\beta(\hat{y},\hat{\bm{x}},\hat{\bm{z}})>0.25$. This also explains the smaller FRL ranges we observe for less extreme values of the Brier score, as they can be observed in the boxplots obtained by equal-frequency binning with respect to the Brier score and depicted on the upper part of the figure. 

The most notable aspect of Figure~\ref{fig:adult} is the set of histograms on the right. These depict predictive accuracy across ten equally sized subgroups of instances, defined by the quantiles of the FRL distribution. The results reveal a clear decline in prediction accuracy as FRL increase. This trend is consistent with the conjectured relationship between accuracy and robustness/fairness introduced in Section~\ref{sec:introduction}. 

\begin{figure}[htp!]
\centering
\subfigure[{\tt Adult}]{
\label{fig:hist_adult}
\begin{tikzpicture}[scale=1.05]
\begin{axis}[asse,legend entries={\tiny{Accuracy},\tiny{Brier Score}},legend pos = north east]
\addplot[area legend,acc2] coordinates{(0.0,0.99) (0.02,0.99)};
\addplot[acc2] coordinates{(0.02,0.98) (0.05,0.98)};
\addplot[acc2] coordinates{(0.05,0.96) (0.1,0.96)};
\addplot[acc2] coordinates{(0.1,0.91) (0.17,0.91)};
\addplot[acc2] coordinates{(0.17,0.85) (0.25,0.85)};
\addplot[acc2] coordinates{(0.25,0.73) (0.32,0.73)};
\addplot[acc2] coordinates{(0.32,0.62) (0.39,0.62)};
\addplot[acc2] coordinates{(0.39,0.5) (0.45,0.5)};
\addplot[acc2] coordinates{(0.45,0.32) (0.54,0.32)};
\addplot[acc2] coordinates{(0.54,0.22) (1.0,0.22)};
\addlegendimage{brier}
\addplot[brier] coordinates{(0.0,0.01) (0.02,0.01)};
\addplot[brier] coordinates{(0.02,0.02) (0.05,0.02)};
\addplot[brier] coordinates{(0.05,0.05) (0.1,0.05)};
\addplot[brier] coordinates{(0.1,0.09) (0.17,0.09)};
\addplot[brier] coordinates{(0.17,0.16) (0.25,0.16)};
\addplot[brier] coordinates{(0.25,0.23) (0.32,0.23)};
\addplot[brier] coordinates{(0.32,0.3) (0.39,0.3)};
\addplot[brier] coordinates{(0.39,0.35) (0.45,0.35)};
\addplot[brier] coordinates{(0.45,0.41) (0.54,0.41)};
\addplot[brier] coordinates{(0.54,0.56) (1.0,0.56)};
\end{axis}
\end{tikzpicture}}
\subfigure[{\tt KDD Census-Income}]{
\label{fig:hist_census}
\begin{tikzpicture}[scale=1.05]
\begin{axis}[asse,legend entries={\tiny{Accuracy},\tiny{Brier Score}},legend pos = north east]
\addplot[area legend,acc2] coordinates{(0.0,0.99) (0.01,0.99)};
\addplot[acc2] coordinates{(0.01,0.99) (0.02,0.99)};
\addplot[acc2] coordinates{(0.02,0.98) (0.05,0.98)};
\addplot[acc2] coordinates{(0.05,0.93) (0.09,0.93)};
\addplot[acc2] coordinates{(0.09,0.92) (0.12,0.92)};
\addplot[acc2] coordinates{(0.12,0.86) (0.25,0.86)};
\addplot[acc2] coordinates{(0.25,0.59) (0.92,0.59)};
\addlegendimage{brier}
\addplot[brier] coordinates{(0.0,0.01) (0.01,0.01)};
\addplot[brier] coordinates{(0.01,0.01) (0.02,0.01)};
\addplot[brier] coordinates{(0.02,0.02) (0.05,0.02)};
\addplot[brier] coordinates{(0.05,0.07) (0.09,0.07)};
\addplot[brier] coordinates{(0.09,0.08) (0.12,0.08)};
\addplot[brier] coordinates{(0.12,0.13) (0.25,0.13)};
\addplot[brier] coordinates{(0.25,0.28) (0.92,0.28)};
\end{axis}
\end{tikzpicture}}
\caption{FRL versus accuracy and Brier score histograms for datasets where at least one private feature remains predictive of the target in each fold.\label{fig:adult_census}}
\end{figure}
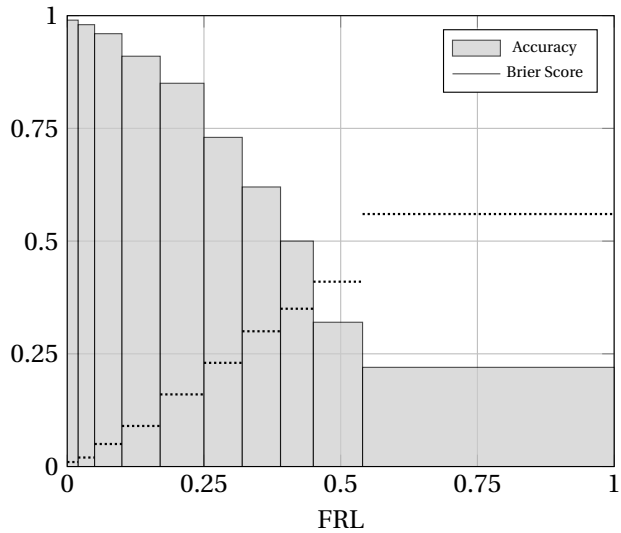
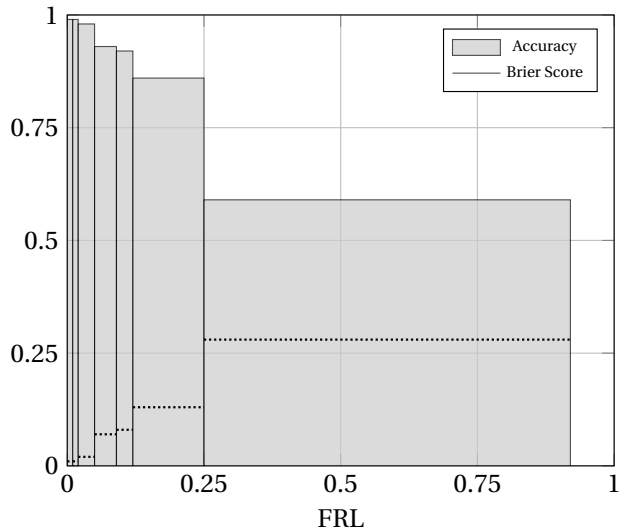

Figure~\ref{fig:hist_adult} presents the same accuracy histograms alongside the corresponding Brier score levels. Predictive performance deteriorates with increasing FRL values, as indicated by both the decline in accuracy and the rise in Brier losses (dashed lines). A similar pattern emerges in Figure~\ref{fig:hist_census}, where the {\tt KDD Census-Income} dataset is considered.

As shown in Table~\ref{tab:datasets}, {\tt Adult} and {\tt KDD Census-Income} are the only datasets of our benchmark not achieving fairness by design on any fold (i.e., such that $k=0$), this meaning that no BN learned from these datasets makes all the private features irrelevant to the target.

\begin{figure}[htp!]
\centering
\subfigure[{\tt Law school}]{
\label{fig:hist_law}
\begin{tikzpicture}[scale=0.7]
\begin{axis}[asse,xlabel=$\mathrm{FRL}$,xmin=-0.02,xmax=0.3,ymax=1.1,legend entries={\tiny{Accuracy ($\mathrm{FRL}>0$)},\tiny{Accuracy ($\mathrm{FRL}=0$)},\tiny{Brier Score}},legend pos = north east]
\addplot[area legend,acc2] coordinates{(0.0,0.99) (0.01,0.99)};
\addplot[area legend,acc1] coordinates{(-0.02,0.99) (0.0,0.99)};
\addlegendimage{brier}
\addplot[brier] coordinates{(-0.02,0.01) (0.0,0.01)};
\addplot[brier] coordinates{(0.0,0.01) (0.01,0.01)};
\addplot[acc2] coordinates{(0.01,0.98) (0.02,0.98)};
\addplot[brier] coordinates{(0.01,0.02) (0.02,0.02)};
\addplot[acc2] coordinates{(0.02,0.95) (0.04,0.95)};
\addplot[brier] coordinates{(0.02,0.05) (0.04,0.05)};
\addplot[acc2] coordinates{(0.04,0.91) (0.09,0.91)};
\addplot[brier] coordinates{(0.04,0.08) (0.09,0.08)};
\addplot[acc2] coordinates{(0.09,0.76) (0.28,0.76)};
\addplot[brier] coordinates{(0.09,0.18) (0.28,0.18)};
\end{axis}
\end{tikzpicture}}
\subfigure[{\tt Bank marketing}]{
\label{fig:hist_marketing}
\begin{tikzpicture}[scale=0.7]
\begin{axis}[asse,xmin=-0.02,legend entries={\tiny{Accuracy ($\mathrm{FRL}>0$)},\tiny{Accuracy ($\mathrm{FRL}=0$)},\tiny{Brier Score}},legend pos = north east]
\addplot[area legend,acc2] coordinates{(0.0,0.9) (0.15,0.9)};
\addplot[area legend,acc1] coordinates{(-0.02,0.9) (0.0,0.9)};
\addplot[brier] coordinates{(-0.02,0.08) (0.0,0.08)};
\addplot[brier] coordinates{(0.0,0.1) (0.15,0.1)};
\addplot[acc2] coordinates{(0.15,0.74) (0.4,0.74)};
\addplot[brier] coordinates{(0.15,0.2) (0.4,0.2)};
\addplot[acc2] coordinates{(0.4,0.51) (0.48,0.51)};
\addplot[brier] coordinates{(0.4,0.26) (0.48,0.26)};
\addplot[acc2] coordinates{(0.48,0.05) (0.53,0.05)};
\addplot[brier] coordinates{(0.48,0.31) (0.53,0.31)};
\addplot[acc2] coordinates{(0.53,0.07) (0.94,0.07)};
\addplot[brier] coordinates{(0.53,0.44) (0.94,0.44)};
\end{axis}
\end{tikzpicture}}
\vskip 10mm
\subfigure[{\tt COMPAS recid.}]{
\label{fig:hist_compas}
\begin{tikzpicture}[scale=0.7]
\begin{axis}[asse,xmin=-0.02,xmax=0.6,ymax=0.75,legend entries={\tiny{Accuracy ($\mathrm{FRL}>0$)},\tiny{Accuracy ($\mathrm{FRL}=0$)},\tiny{Brier Score}},legend pos = north east]
\addplot[area legend,acc2] coordinates{(0.0,0.66) (0.0,0.66)};
\addplot[area legend,acc1] coordinates{(-0.02,0.64) (0.0,0.64)};
\addlegendimage{brier}
\addplot[acc2] coordinates{(0.0,0.72) (0.06,0.72)};
\addplot[acc2] coordinates{(0.06,0.67) (0.08,0.67)};
\addplot[acc2] coordinates{(0.08,0.68) (0.08,0.68)};
\addplot[acc2] coordinates{(0.08,0.69) (0.08,0.69)};
\addplot[acc2] coordinates{(0.08,0.69) (0.09,0.69)};
\addplot[acc2] coordinates{(0.09,0.59) (0.1,0.59)};
\addplot[acc2] coordinates{(0.1,0.53) (0.11,0.53)};
\addplot[acc2] coordinates{(0.11,0.52) (0.56,0.52)};
\addplot[brier] coordinates{(-0.02,0.22) (0.0,0.22)};
\addplot[brier] coordinates{(0.0,0.22) (0.0,0.22)};
\addplot[brier] coordinates{(0.0,0.2) (0.06,0.2)};
\addplot[brier] coordinates{(0.06,0.21) (0.08,0.21)};
\addplot[brier] coordinates{(0.08,0.23) (0.08,0.23)};
\addplot[brier] coordinates{(0.08,0.22) (0.08,0.22)};
\addplot[brier] coordinates{(0.08,0.22) (0.09,0.22)};
\addplot[brier] coordinates{(0.09,0.24) (0.1,0.24)};
\addplot[brier] coordinates{(0.1,0.25) (0.11,0.25)};
\addplot[brier] coordinates{(0.11,0.27) (0.56,0.27)};
\end{axis}
\end{tikzpicture}}
\subfigure[{\tt COMPAS violent recid.}]{
\begin{tikzpicture}[scale=0.7]
\begin{axis}[asse,xmin=-0.02,xmax=0.3,legend entries={\tiny{Accuracy ($\mathrm{FRL}>0$)},\tiny{Accuracy ($\mathrm{FRL}=0$)},\tiny{Brier Score}},legend pos = north east]
\addplot[area legend,acc2] coordinates{(0.0,0.87) (0.041,0.87)};
\addplot[area legend,acc1] coordinates{(-0.02,0.84) (0.0,0.84)};
\addlegendimage{brier}
\addplot[acc2] coordinates{(0.041,0.91) (0.047,0.91)};
\addplot[acc2] coordinates{(0.047,0.91) (0.05,0.91)};
\addplot[acc2] coordinates{(0.05,0.91) (0.053,0.91)};
\addplot[acc2] coordinates{(0.053,0.83) (0.067,0.83)};
\addplot[acc2] coordinates{(0.067,0.81) (0.098,0.81)};
\addplot[acc2] coordinates{(0.098,0.62) (0.294,0.62)};
\addplot[brier] coordinates{(-0.02,0.12) (0.0,0.12)};
\addplot[brier] coordinates{(0.0,0.1) (0.041,0.1)};
\addplot[brier] coordinates{(0.041,0.08) (0.047,0.08)};
\addplot[brier] coordinates{(0.047,0.08) (0.05,0.08)};
\addplot[brier] coordinates{(0.05,0.08) (0.053,0.08)};
\addplot[brier] coordinates{(0.053,0.14) (0.067,0.14)};
\addplot[brier] coordinates{(0.067,0.16) (0.098,0.16)};
\addplot[brier] coordinates{(0.098,0.25) (0.294,0.25)};
\end{axis}
\end{tikzpicture}}
\caption{FRL versus accuracy and Brier score histograms for datasets where at least one private feature remains predictive of the target in at least a fold.}\label{fig:extra}
\end{figure}
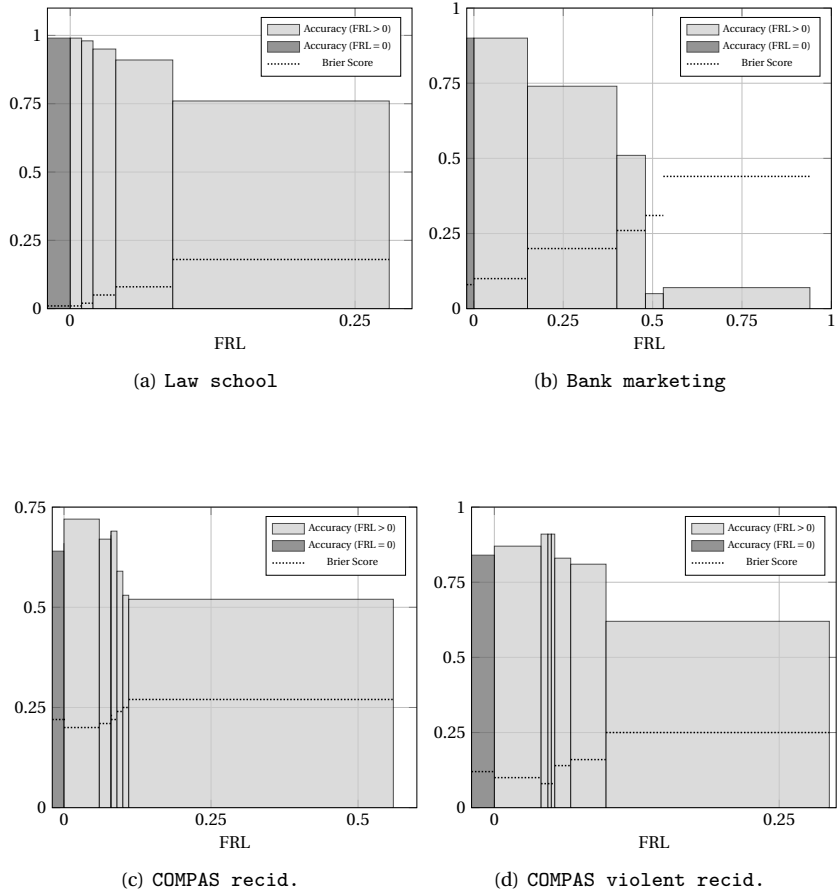

Let us now consider the datasets for which fairness by design is achieved only on some folds, but not on all of them (i.e., $0<k<10$). On these folds we have FRL$=0$ by construction. For an effective visualisation of the relation between FRL and the predictive accuracy (or Brier score) we use the same kind of histograms as in Figure~\ref{fig:adult_census}, with an additional bar/line of fixed width on the left side, depicting the predictive performance on the instances with FRL$=0$. The results are in Figure~\ref{fig:extra}. Also for these models we observe a clear degradation of the predictive performance for increasing FRL levels.

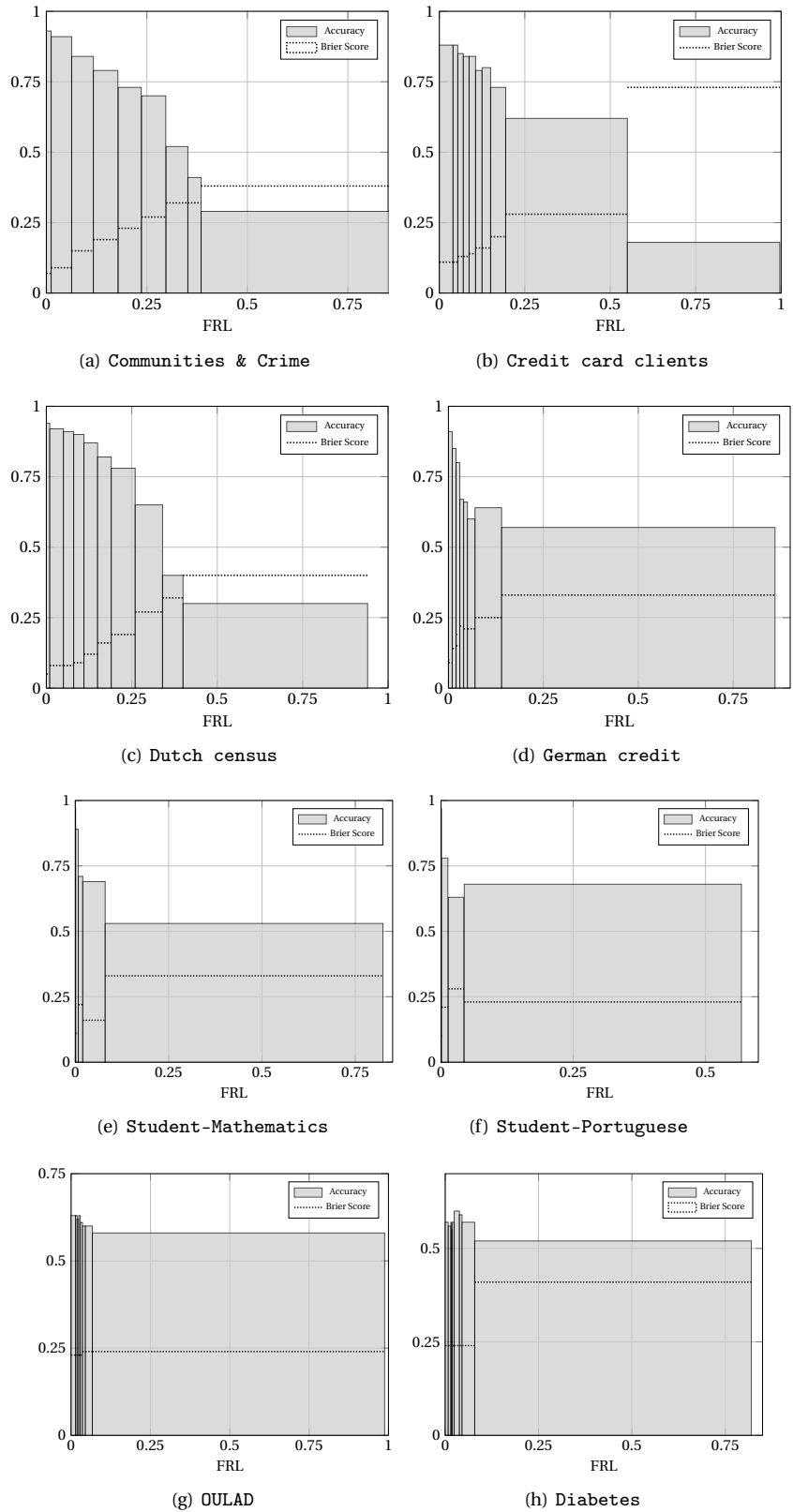
\begin{figure}[htp!]
\centering
\subfigure[{\tt Communities \& Crime}]{
\begin{tikzpicture}[scale=0.7]
\begin{axis}[asse,xmax=0.85,legend entries={\tiny{Accuracy},\tiny{Brier Score}},legend pos=north east]
\addplot[area legend,acc2] coordinates{(0.0,0.99) (0.0,0.99)};
\addplot[area legend,brier] coordinates{(0.0,0.01) (0.0,0.01)};
\addplot[acc2] coordinates{(0.0,0.93) (0.012,0.93)};
\addplot[brier] coordinates{(0.0,0.07) (0.012,0.07)};
\addplot[acc2] coordinates{(0.012,0.91) (0.063,0.91)};
\addplot[brier] coordinates{(0.012,0.09) (0.063,0.09)};
\addplot[acc2] coordinates{(0.063,0.84) (0.117,0.84)};
\addplot[brier] coordinates{(0.063,0.15) (0.117,0.15)};
\addplot[acc2] coordinates{(0.117,0.79) (0.178,0.79)};
\addplot[brier] coordinates{(0.117,0.19) (0.178,0.19)};
\addplot[acc2] coordinates{(0.179,0.73) (0.236,0.73)};
\addplot[brier] coordinates{(0.179,0.23) (0.236,0.23)};
\addplot[acc2] coordinates{(0.237,0.7) (0.297,0.7)};
\addplot[brier] coordinates{(0.237,0.27) (0.297,0.27)};
\addplot[acc2] coordinates{(0.298,0.52) (0.352,0.52)};
\addplot[brier] coordinates{(0.298,0.32) (0.352,0.32)};
\addplot[acc2] coordinates{(0.352,0.41) (0.385,0.41)};
\addplot[brier] coordinates{(0.352,0.32) (0.385,0.32)};
\addplot[acc2] coordinates{(0.385,0.29) (0.853,0.29)};
\addplot[brier] coordinates{(0.385,0.38) (0.853,0.38)};
\end{axis}
\end{tikzpicture}}
\subfigure[{\tt Credit card clients}]{
\begin{tikzpicture}[scale=0.7]
\begin{axis}[asse,legend entries={\tiny{Accuracy},\tiny{Brier Score}},legend pos=north east]
\addplot[area legend,acc2] coordinates{(0.0,0.88) (0.04,0.88)};
\addlegendimage{brier}
\addplot[acc2] coordinates{(0.04,0.88) (0.054,0.88)};
\addplot[acc2] coordinates{(0.054,0.85) (0.07,0.85)};
\addplot[acc2] coordinates{(0.07,0.84) (0.087,0.84)};
\addplot[acc2] coordinates{(0.087,0.84) (0.106,0.84)};
\addplot[acc2] coordinates{(0.106,0.79) (0.125,0.79)};
\addplot[acc2] coordinates{(0.125,0.8) (0.15,0.8)};
\addplot[acc2] coordinates{(0.15,0.73) (0.194,0.73)};
\addplot[acc2] coordinates{(0.194,0.62) (0.55,0.62)};
\addplot[acc2] coordinates{(0.55,0.18) (0.996,0.18)};
\addplot[brier] coordinates{(0.0,0.11) (0.04,0.11)};
\addplot[brier] coordinates{(0.04,0.11) (0.054,0.11)};
\addplot[brier] coordinates{(0.054,0.13) (0.07,0.13)};
\addplot[brier] coordinates{(0.07,0.13) (0.087,0.13)};
\addplot[brier] coordinates{(0.087,0.14) (0.106,0.14)};
\addplot[brier] coordinates{(0.106,0.16) (0.125,0.16)};
\addplot[brier] coordinates{(0.125,0.16) (0.15,0.16)};
\addplot[brier] coordinates{(0.15,0.2) (0.194,0.2)};
\addplot[brier] coordinates{(0.194,0.28) (0.55,0.28)};
\addplot[brier] coordinates{(0.55,0.73) (0.996,0.73)};
\end{axis}
\end{tikzpicture}}
\subfigure[{\tt Dutch census}]{
\begin{tikzpicture}[scale=0.7]
\begin{axis}[asse,xlabel=$\mathrm{FRL}$,legend entries={\tiny{Accuracy},\tiny{Brier Score}},legend pos=north east]
\addplot[area legend,acc2] coordinates{(0.0,0.94) (0.01,0.94)};
\addlegendimage{brier}
\addplot[acc2] coordinates{(0.01,0.92) (0.05,0.92)};
\addplot[acc2] coordinates{(0.05,0.91) (0.08,0.91)};
\addplot[acc2] coordinates{(0.08,0.9) (0.11,0.9)};
\addplot[acc2] coordinates{(0.11,0.87) (0.15,0.87)};
\addplot[acc2] coordinates{(0.15,0.82) (0.19,0.82)};
\addplot[acc2] coordinates{(0.19,0.78) (0.26,0.78)};
\addplot[acc2] coordinates{(0.26,0.65) (0.34,0.65)};
\addplot[acc2] coordinates{(0.34,0.4) (0.4,0.4)};
\addplot[acc2] coordinates{(0.4,0.3) (0.94,0.3)};
\addplot[brier] coordinates{(0.0,0.05) (0.01,0.05)};
\addplot[brier] coordinates{(0.01,0.08) (0.05,0.08)};
\addplot[brier] coordinates{(0.05,0.08) (0.08,0.08)};
\addplot[brier] coordinates{(0.08,0.09) (0.11,0.09)};
\addplot[brier] coordinates{(0.11,0.12) (0.15,0.12)};
\addplot[brier] coordinates{(0.15,0.16) (0.19,0.16)};
\addplot[brier] coordinates{(0.19,0.19) (0.26,0.19)};
\addplot[brier] coordinates{(0.26,0.27) (0.34,0.27)};
\addplot[brier] coordinates{(0.34,0.32) (0.4,0.32)};
\addplot[brier] coordinates{(0.4,0.4) (0.94,0.4)};
\end{axis}
\end{tikzpicture}}
\subfigure[{\tt German credit}]{
\begin{tikzpicture}[scale=0.7]
\begin{axis}[asse,xmax=0.9,xlabel=$\mathrm{FRL}$,legend entries={\tiny{Accuracy},\tiny{Brier Score}},legend pos=north east]
\addplot[area legend,acc2] coordinates{(0.0,0.9) (0.0,0.9)};
\addlegendimage{brier}
\addplot[acc2] coordinates{(0.0,0.91) (0.01,0.91)};
\addplot[acc2] coordinates{(0.01,0.85) (0.02,0.85)};
\addplot[acc2] coordinates{(0.02,0.77) (0.02,0.77)};
\addplot[acc2] coordinates{(0.02,0.8) (0.03,0.8)};
\addplot[acc2] coordinates{(0.03,0.67) (0.04,0.67)};
\addplot[acc2] coordinates{(0.04,0.66) (0.05,0.66)};
\addplot[acc2] coordinates{(0.05,0.6) (0.07,0.6)};
\addplot[acc2] coordinates{(0.07,0.64) (0.14,0.64)};
\addplot[acc2] coordinates{(0.14,0.57) (0.86,0.57)};
\addplot[brier] coordinates{(0.0,0.1) (0.0,0.1)};
\addplot[brier] coordinates{(0.0,0.09) (0.01,0.09)};
\addplot[brier] coordinates{(0.01,0.14) (0.02,0.14)};
\addplot[brier] coordinates{(0.02,0.19) (0.02,0.19)};
\addplot[brier] coordinates{(0.02,0.15) (0.03,0.15)};
\addplot[brier] coordinates{(0.03,0.22) (0.04,0.22)};
\addplot[brier] coordinates{(0.04,0.21) (0.05,0.21)};
\addplot[brier] coordinates{(0.05,0.21) (0.07,0.21)};
\addplot[brier] coordinates{(0.07,0.25) (0.14,0.25)};
\addplot[brier] coordinates{(0.14,0.33) (0.86,0.33)};
\end{axis}
\end{tikzpicture}}
\subfigure[{\tt Student-Mathematics}]{
\begin{tikzpicture}[scale=0.65]
\begin{axis}[asse,xmax=0.85,legend entries={\tiny{Accuracy},\tiny{Brier Score}},legend pos=north east]
\addplot[area legend,acc2] coordinates{(0.0,0.89) (0.0,0.89)};
\addlegendimage{brier}
\addplot[acc2] coordinates{(0.0,0.97) (0.001,0.97)};
\addplot[acc2] coordinates{(0.001,0.89) (0.008,0.89)};
\addplot[acc2] coordinates{(0.008,0.71) (0.02,0.71)};
\addplot[acc2] coordinates{(0.02,0.69) (0.08,0.69)};
\addplot[acc2] coordinates{(0.08,0.53) (0.824,0.53)};
\addplot[brier] coordinates{(0.0,0.03) (0.001,0.03)};
\addplot[brier] coordinates{(0.001,0.11) (0.008,0.11)};
\addplot[brier] coordinates{(0.008,0.22) (0.02,0.22)};
\addplot[brier] coordinates{(0.02,0.16) (0.08,0.16)};
\addplot[brier] coordinates{(0.08,0.33) (0.824,0.33)};
\end{axis}
\end{tikzpicture}\label{fig:hist_math}}
\subfigure[{\tt Student-Portuguese}]{
\begin{tikzpicture}[scale=0.65]
\begin{axis}[asse,xmax=0.6,legend entries={\tiny{Accuracy},\tiny{Brier Score}},legend pos=north east]
\addplot[area legend,acc2] coordinates{(0.0,0.87) (0.0,0.87)};
\addlegendimage{brier}
\addplot[acc2] coordinates{(0.0,1.0) (0.0,1.0)};
\addplot[acc2] coordinates{(0.0,1.0) (0.0,1.0)};
\addplot[acc2] coordinates{(0.0,1.0) (0.0,1.0)};
\addplot[acc2] coordinates{(0.0,1.0) (0.0,1.0)};
\addplot[acc2] coordinates{(0.0,1.0) (0.0,1.0)};
\addplot[acc2] coordinates{(0.0,0.97) (0.001,0.97)};
\addplot[acc2] coordinates{(0.001,0.78) (0.013,0.78)};
\addplot[acc2] coordinates{(0.014,0.63) (0.043,0.63)};
\addplot[acc2] coordinates{(0.044,0.68) (0.568,0.68)};
\addplot[brier] coordinates{(0.0,0.1) (0.0,0.1)};
\addplot[brier] coordinates{(0.0,0.0) (0.0,0.0)};
\addplot[brier] coordinates{(0.0,0.0) (0.0,0.0)};
\addplot[brier] coordinates{(0.0,0.0) (0.0,0.0)};
\addplot[brier] coordinates{(0.0,0.0) (0.0,0.0)};
\addplot[brier] coordinates{(0.0,0.0) (0.0,0.0)};
\addplot[brier] coordinates{(0.0,0.03) (0.001,0.03)};
\addplot[brier] coordinates{(0.001,0.21) (0.013,0.21)};
\addplot[brier] coordinates{(0.014,0.28) (0.043,0.28)};
\addplot[brier] coordinates{(0.044,0.23) (0.568,0.23)};
\end{axis}
\end{tikzpicture}}
\subfigure[{\tt OULAD}]{
\begin{tikzpicture}[scale=0.65]
\begin{axis}[asse,ymax=0.75,legend entries={\tiny{Accuracy},\tiny{Brier Score}},legend pos=north east]
\addplot[area legend,acc2] coordinates{(0.0,0.63) (0.014,0.63)};
\addlegendimage{brier}
\addplot[brier] coordinates{(0.0,0.23) (0.014,0.23)};
\addplot[acc2] coordinates{(0.014,0.63) (0.019,0.63)};
\addplot[brier] coordinates{(0.014,0.23) (0.019,0.23)};
\addplot[acc2] coordinates{(0.019,0.62) (0.024,0.62)};
\addplot[brier] coordinates{(0.019,0.23) (0.024,0.23)};
\addplot[acc2] coordinates{(0.024,0.63) (0.029,0.63)};
\addplot[brier] coordinates{(0.024,0.23) (0.029,0.23)};
\addplot[acc2] coordinates{(0.029,0.61) (0.036,0.61)};
\addplot[brier] coordinates{(0.029,0.23) (0.036,0.23)};
\addplot[acc2] coordinates{(0.036,0.6) (0.045,0.6)};
\addplot[brier] coordinates{(0.036,0.24) (0.045,0.24)};
\addplot[acc2] coordinates{(0.045,0.6) (0.067,0.6)};
\addplot[brier] coordinates{(0.045,0.24) (0.067,0.24)};
\addplot[acc2] coordinates{(0.067,0.58) (0.988,0.58)};
\addplot[brier] coordinates{(0.067,0.24) (0.988,0.24)};
\end{axis}
\end{tikzpicture}}
\subfigure[{\tt Diabetes}]{
\begin{tikzpicture}[scale=0.65]
\begin{axis}[asse,xmax=0.85,ymax=0.7,legend entries={\tiny{Accuracy},\tiny{Brier Score}},legend pos = north east]
\addplot[area legend,acc2] coordinates{(0.0,0.57) (0.008,0.57)};
\addplot[area legend,brier] coordinates{(0.0,0.24) (0.008,0.24)};
\addplot[acc2] coordinates{(0.008,0.56) (0.015,0.56)};
\addplot[brier] coordinates{(0.008,0.24) (0.015,0.24)};
\addplot[acc2] coordinates{(0.015,0.57) (0.017,0.57)};
\addplot[brier] coordinates{(0.015,0.24) (0.017,0.24)};
\addplot[acc2] coordinates{(0.017,0.55) (0.019,0.55)};
\addplot[brier] coordinates{(0.017,0.25) (0.019,0.25)};
\addplot[acc2] coordinates{(0.019,0.57) (0.024,0.57)};
\addplot[brier] coordinates{(0.019,0.24) (0.024,0.24)};
\addplot[acc2] coordinates{(0.024,0.6) (0.038,0.6)};
\addplot[brier] coordinates{(0.024,0.24) (0.038,0.24)};
\addplot[acc2] coordinates{(0.038,0.59) (0.045,0.59)};
\addplot[brier] coordinates{(0.038,0.24) (0.045,0.24)};
\addplot[acc2] coordinates{(0.045,0.57) (0.079,0.57)};
\addplot[brier] coordinates{(0.045,0.24) (0.079,0.24)};
\addplot[acc2] coordinates{(0.079,0.52) (0.82,0.52)};
\addplot[brier] coordinates{(0.079,0.41) (0.82,0.41)};
\end{axis}
\end{tikzpicture}\label{fig:diabetes}}
\caption{FRL vs. accuracy/Brier for datasets where the target is forced to be a parent of the features.}\label{fig:forced}
\end{figure}

On the remaining eight datasets, fairness is achieved by design on each fold (i.e., $k=10$, see Table~\ref{tab:datasets}) and we have zero FRL on all the instances in the dataset. Yet, for a further validation of our conjecture, we re-train the BN structures by forcing arcs between the target and all the features. This can be trivially done by constraining the local search meta-heuristics we adopt. The results, depicted in Figure~\ref{fig:forced}, confirm the trend observed in the previous experiments. We notice that degradation of the performance for higher FRLs is typically less evident on datasets on which the predictive performance of the BNs is poorer (such as those in Figs.~\ref{fig:hist_math}-\ref{fig:diabetes}).

\begin{figure}[htp!]
\centering
\begin{tikzpicture}[scale=0.95]
\begin{axis}
[boxplot/draw direction = y,
xlabel near ticks,
ylabel near ticks,
ylabel = {$\frac{T_{\mathrm{FRL-BN}}}{T_{\mathrm{FRL-MRF}}}$},
ylabel style = {rotate=-90},
xtick = {1, 2, 3, 4, 5, 6, 7, 8, 9, 10, 11, 12, 13, 14},
ytick = {1, 3, 5},
xticklabel style = {align=center, font=\tiny, rotate=70},
xtick style = {draw=none}, 
xlabel style={font=\tiny},
xticklabels = {
    {\tt Adult} (3), 
    {\tt Bank Marketing} (2), 
    {\tt Communities \& Crime} (1), 
    {\tt COMPAS recid.} (2),
    {\tt COMPAS violent recid.} (2),
    {\tt Credit card clients} (3), 
    {\tt Diabetes} (1), 
    {\tt Dutch Census} (1), 
    {\tt German Credit} (2),
    {\tt KDD Census-Income} (2), 
    {\tt Law school} (2),
    {\tt OULAD} (1),
    {\tt Student-Mathematics} (2), 
    {\tt Student-Portuguese} (2), 
    },
every axis plot/.append style = {},]
\addplot+[mark=none,black!20,thick] coordinates {(0,1) (14,1)};
\addplot + [mark = *,boxplot,mark options={black},color=black,fill=black,draw=black,fill opacity=.1] table[col sep=comma,y index=3]{Ratios/adult.dat};
\addplot + [mark = *,boxplot,mark options={black},color=black,fill=black,draw=black,fill opacity=.1] table[col sep=comma,y index=3]{Ratios/marketing.dat};
\addplot + [mark = *,boxplot,mark options={black},color=black,fill=black,draw=black,fill opacity=.1] table[col sep=comma,y index=3]{Ratios/crime.dat};
\addplot + [mark = *,boxplot,mark options={black},color=black,fill=black,draw=black,fill opacity=.1] table[col sep=comma,y index=3]{Ratios/compas.dat};
\addplot + [mark = *,boxplot,mark options={black},color=black,fill=black,draw=black,fill opacity=.1] table[col sep=comma,y index=3]{Ratios/violent.dat};
\addplot + [mark = *,boxplot,mark options={black},color=black,fill=black,draw=black,fill opacity=.1] table[col sep=comma,y index=3]{Ratios/credit.dat};
\addplot + [mark = *,boxplot,mark options={black},color=black,fill=black,draw=black,fill opacity=.1] table[col sep=comma,y index=2]{Ratios/diabetes.dat};
\addplot + [mark = *,boxplot,mark options={black},color=black,fill=black,draw=black,fill opacity=.1] table[col sep=comma,y index=3]{Ratios/dutch.dat};
\addplot + [mark = *,boxplot,mark options={black},color=black,fill=black,draw=black,fill opacity=.1] table[col sep=comma,y index=3]{Ratios/german.dat};
\addplot + [mark = *,boxplot,mark options={black},color=black,fill=black,draw=black,fill opacity=.1] table[col sep=comma,y index=3]{Ratios/census.dat};
\addplot + [mark = *,boxplot,mark options={black},color=black,fill=black,draw=black,fill opacity=.1] table[col sep=comma,y index=3]{Ratios/law.dat};
\addplot + [mark = *,boxplot,mark options={black},color=black,fill=black,draw=black,fill opacity=.1] table[col sep=comma,y index=2]{Ratios/oulad.dat};
\addplot + [mark = *,boxplot,mark options={black},color=black,fill=black,draw=black,fill opacity=.1] table[col sep=comma,y index=3]{Ratios/student-mat.dat};
\addplot + [mark = *,boxplot,mark options={black},color=black,fill=black,draw=black,fill opacity=.1] table[col sep=comma,y index=3]{Ratios/student-por.dat};
\end{axis}
\end{tikzpicture}
\caption{Relative execution times for FRL computation in the BN (Eq.~\eqref{eq:frl}) with respect to the MRF approach (Section~\ref{sec:mrf}). The x-labels denote the dataset and the number of private features.\label{fig:speeds}}
\end{figure}
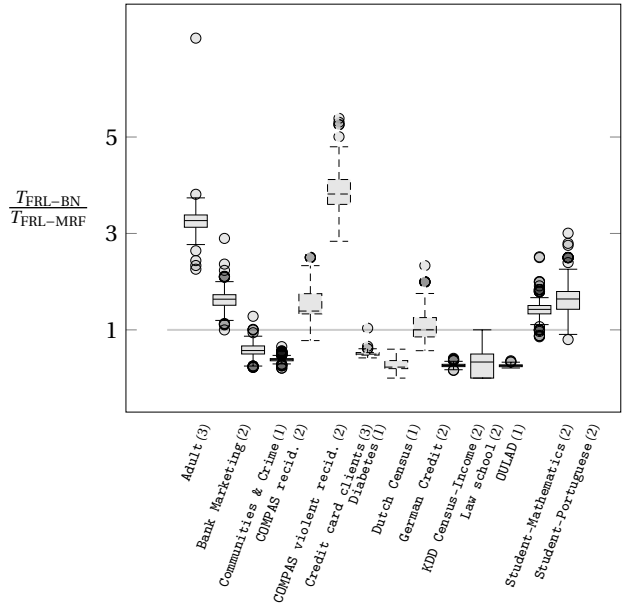

Finally, to evaluate the computational advantages of the MRF-based approach to the computation of the FRL with respect to the brute-force BN method, Figure~\ref{fig:speeds} shows boxplots of the ratio between the execution times for each test instance across the two methods. The low number of private features (i.e., $|\bm{X}|\leq 3$) makes the difference in performance between the two approaches small. Nevertheless, as expected, the datasets with three private features exhibits the largest discrepancy. 

\begin{figure}[htp!]
\centering
\begin{tikzpicture}[scale=0.95]
\begin{axis}
[xlabel={$|\bm{X}|$},
boxplot/draw direction = y,
ymode=log,
log ticks with fixed point,
ylabel = {$\frac{T_{\mathrm{FRL-BN}}}{T_{\mathrm{FRL-MRF}}}$},
ylabel style = {rotate=-90},
xlabel near ticks,
ylabel near ticks,
xtick = {1,2,3,4,5,6,7,8,9,10},
ytick = {1, 10, 100, 300},
xticklabel style = {align=center, font=\scriptsize},
xtick style = {draw=none}, 
xlabel style={font=\tiny},
xticklabels = {2,3,4,5,6,7,8,9,10,11},
every axis plot/.append style = {},]
\addplot + [mark = *,boxplot,mark options = {black},color=black, fill=black, draw=black,fill opacity = .1] table[col sep=comma,y index=1]{./Ratios/ratios1.dat};
\addplot + [mark = *,boxplot,mark options = {black},color=black, fill=black, draw=black,fill opacity = .1] table[col sep=comma,y index=2]{./Ratios/ratios1.dat};
\addplot + [mark = *,boxplot,mark options = {black},color=black, fill=black, draw=black,fill opacity = .1] table[col sep=comma,y index=3]{./Ratios/ratios1.dat};
\addplot + [mark = *,boxplot,mark options = {black},color=black, fill=black, draw=black,fill opacity = .1] table[col sep=comma,y index=4]{./Ratios/ratios1.dat};
\addplot + [mark = *,boxplot,mark options = {black},color=black, fill=black, draw=black,fill opacity = .1] table[col sep=comma,y index=5]{./Ratios/ratios1.dat};
\addplot + [mark = *,boxplot,mark options = {black},color=black, fill=black, draw=black,fill opacity = .1] table[col sep=comma,y index=6]{./Ratios/ratios1.dat};
\addplot + [mark = *,boxplot,mark options = {black},color=black, fill=black, draw=black,fill opacity = .1] table[col sep=comma,y index=7]{./Ratios/ratios1.dat};
\addplot + [mark = *,boxplot,mark options = {black},color=black, fill=black, draw=black,fill opacity = .1] table[col sep=comma,y index=8]{./Ratios/ratios1.dat};
\addplot + [mark = *,boxplot,mark options = {black},color=black, fill=black, draw=black,fill opacity = .1] table[col sep=comma,y index=9]{./Ratios/ratios1.dat};
\end{axis}
\end{tikzpicture}
\caption{Relative execution times for FRL computation in the BN (Eq.~\eqref{eq:frl}) with respect to the MRF approach (Section~\ref{sec:mrf}) for in increasing number of features treated as private in the {\tt Student-Mathematics} dataset.\label{fig:speed}}
\end{figure}
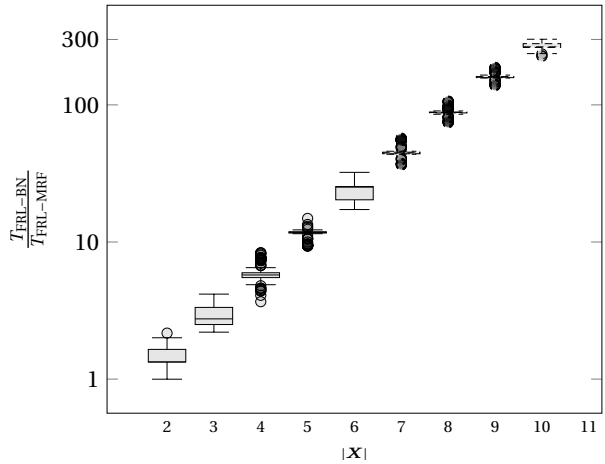

Finally, Figure~\ref{fig:speed} presents the boxplots of the same ratio while treating as private an increasing number of public features of the {\tt Student-Mathematics} dataset. These results underscore the computational efficiency of the MRF-based method, particularly as the number of private features grows.

\section{Limitations and Outlooks}\label{sec:limitations}
The analysis conducted in this study is purely \emph{correlational}, and the directed graphical models employed here are not intended to capture causal relationships. Accordingly, the fairness deviation measure we derived via the FRL does not carry a causal interpretation, and the different instantiations of the private features should not be interpreted as interventions. The approach of Ple\v{c}ko and Bareinboim \cite{plecko2024causal} on \emph{counterfactual fairness} offers a promising foundation for integrating causal reasoning into the models discussed here.

Our work focuses on discrete features for the sake of a simple presentation. An extension to a hybrid setup mixing discrete and continuous features could be obtained within the framework of linear-Gaussian BNs \cite{koller2009}, for which also extensions to the causal setup have been already discussed.  The situation is different for the target variable: extending the framework to regression tasks with a continuous target would require a substantial reformulation of the concepts introduced in this paper. In contrast, our focus on binary classes can be naturally extended to non-binary targets by conducting classification and FRL evaluations through pairwise dominance tests applied to each pair of classes.

Regarding the measure considered to evaluate the robustness of the posterior inferences, the current focus on the Manhattan distance has clear computational advantages related to the linear structure of such measure. Coping with non-linear alternatives (such as the KL) could be much more challenging, while an extension to variance and similar measure of variation could be more manageable, and naturally allow for the extension to continuous targets.

Once our framework will be extended to hybrid models that incorporate both continuous and discrete variables, the experimental evaluation should be revisited. A more refined representation of the model variables is expected to enhance the overall predictive accuracy. Assessing whether the discriminative power of the FRL remains stable under these conditions constitutes a necessary future investigation.
A validation under \emph{controlled bias} scenarios\footnote{See, e.g.,
\href{https://github.com/rcrupiISP/BiasOnDemand}{github.com/rcrupiISP/BiasOnDemand}.} should be also considered.

Finally, extending the framework from individual to \emph{group} fairness would be conceptually straightforward, although it would preclude the use of the MRF mapping. To maintain computational tractability and avoid costly inference procedures, more efficient models—such as probabilistic circuits \cite{choi2020probabilistic}—could serve as practical alternatives to BNs.

\section{Conclusions}\label{sec:conclusions}
We examined whether the correlation between robustness and predictive accuracy — commonly observed in the field of robust probabilistic models — also holds in a  fairness-aware setting. The results of an extensive empirical evaluation are promising: predictive accuracy consistently declines for less robust instances. This observation opens new avenues for mitigating the classical fairness–accuracy trade-off. Specifically, instances with high fairness scores can be handled by standard models, while a fairness-constrained, potentially less accurate model could be selectively applied to non-robust instances. Exploring this targeted approach to fairness mitigation represents an exciting and necessary direction for future research in machine learning.

\section*{Acknowledgements}
The research of Eric Rossetto was funded by Swiss Post AG as part of a Doctoral Studies Grant on \emph{AI and Robustness}.

\newpage
\appendix
\section{Proofs}
In this appendix, we gather the proofs of the two technical results of the paper.
\begin{proofprop}
The following straightforward result: 
\begin{equation}
\max_t \left|f(t)\right|= \max \left\{ \max_t \left[ f(t) \right] , \max_t \left[ -f(t) \right] \right\}   \,,
\end{equation}
allows to rewrite the maximum of the optimisation in Eq.~\eqref{eq:max} as follows:
\begin{equation}
\max \left\{ \max_{\bm{x}\in\Omega_{\bm{X}}} \left[ P(y_0|\bm{x},\hat{\bm{z}})-P(y_0|\hat{\bm{x}},\hat{\bm{z}}) \right] \,, \max_{\bm{x}\in\Omega_{\bm{X}}} \left[ P(y_0|\hat{\bm{x}},\hat{\bm{z}})-P(y_0|\bm{x},\hat{\bm{z}})\right]\right\} 
\end{equation}
As $P(y_0|\hat{\bm{x}},\hat{\bm{z}})$ is a constant, we can move the maximisation with respect to $\bm{x}$ and obtain the maximum and the minimum in Eqs.~\eqref{eq:cir1} and \eqref{eq:cir2}, i.e.,
\begin{equation}
\max \left\{  \left[ P(y_0|\overline{\bm{x}},\hat{\bm{z}})-P(y_0|\hat{\bm{x}},\hat{\bm{z}}) \right] \,,  \left[ P(y_0|\hat{\bm{x}},\hat{\bm{z}})-P(y_0|\underline{\bm{x}},\hat{\bm{z}})\right]\right\} \,.
\end{equation}
\end{proofprop}
The above maximum between two quantities can be clearly rewritten as:
\begin{equation}
\left\{
\begin{array}{ll}P(y_0|\overline{\bm{x}},\hat{\bm{z}})-P(y_0|\hat{\bm{x}},\hat{\bm{z}}) & \mathrm{if}\, 
P(y_0|\overline{\bm{x}},\hat{\bm{z}})-P(y_0|\hat{\bm{x}},\hat{\bm{z}})>
P(y_0|\hat{\bm{x}},\hat{\bm{z}})-P(y_0|\underline{\bm{x}},\hat{\bm{z}})\,,
\\ P(y_0|\hat{\bm{x}},\hat{\bm{z}})-P(y_0|\underline{\bm{x}},\hat{\bm{z}}) & \mathrm{otherwise}\,. \end{array}
\right.
\end{equation}
And hence:
\begin{equation}
\left\{
\begin{array}{ll}P(y_0|\overline{\bm{x}},\hat{\bm{z}})-P(y_0|\hat{\bm{x}},\hat{\bm{z}}) & \mathrm{if}\, 
P(y_0|\overline{\bm{x}},\hat{\bm{z}})+
P(y_0|\underline{\bm{x}},\hat{\bm{z}})>2 P(y_0|\hat{\bm{x}},\hat{\bm{z}})\,,
\\ P(y_0|\hat{\bm{x}},\hat{\bm{z}})-P(y_0|\underline{\bm{x}},\hat{\bm{z}}) & \mathrm{otherwise} \end{array}
\right.\,,
\end{equation}
which can be seen as an equivalent formulation of Eq.~\eqref{eq:xstar}, where the values of the maxima appear instead of the points where these maxima are achieved.

\begin{proof}
As we cope with a Boolean target variable, i.e., $|\Omega_{Y}|=2$, the objective function of the optimisations in Eqs.~\eqref{eq:cir1} and \eqref{eq:cir2} rewrites as follows:
\begin{equation}
P(y_0|\bm{x},\hat{\bm{z}})=
\frac{P(y_0,\bm{x},\hat{\bm{z}})}{P(\bm{x},\hat{\bm{z}})}=
\frac{P(y_0,\bm{x},\hat{\bm{z}})}{P(y_0,\bm{x},\hat{\bm{z}})+P(y_1,\bm{x},\hat{\bm{z}})}=
\frac{1}{1+\frac{P(y_1,\bm{x},\hat{\bm{z}})}{P(y_0,\bm{x},\hat{\bm{z}})}}
\end{equation}
As $f(t)=(1+t)^{-1}$ is a monotone decreasing function of $t$, we can address Eq.~\eqref{eq:cir2} as follows:
\begin{equation}\label{eq:ratio}
\overline{\bm{x}}:=\arg \max_{\bm{x} \in \Omega_{\bm{X}}} \frac{P(y_1,\bm{x},\hat{\bm{z}})}{P(y_0,\bm{x},\hat{\bm{z}})}\,.
\end{equation}
The BN factorisation in Eq.~\eqref{eq:joint} allows to rewrite the objective function in Eq.~\eqref{eq:ratio} as follows:
\begin{equation}\label{eq:prods}
\frac{P(y_1,\bm{x},\hat{\bm{z}})}{P(y_0,\bm{x},\hat{\bm{z}})}=
\frac{\prod_{V\in(Y,\bm{X},\bm{Z})} P_1(v|\mathrm{pa}_v)}{\prod_{V\in(Y,\bm{X},\bm{Z})} P_0(v|\mathrm{pa}_V)}
=\prod_{V\in(Y,\bm{X},\bm{Z})} \frac{P_1(v|\mathrm{pa}_v)}{P_0(v|\mathrm{pa}_V)}\,,
\end{equation}
where $P_1$ denotes the elements of the BN CPTs when the states $(v,\mathrm{pa}_V)$ are those consistent with $(y_1,\bm{x},\hat{\bm{z}})$ and $P_0$ when consistent with 
$(y_0,\bm{x},\hat{\bm{z}})$. The product in the rightmost hand side of Eq.~\eqref{eq:prods} can be restricted to ratios of CPT values including $Y$, i.e., the CPTs associated with $Y$ and its children. This corresponds to the joint PMF of the MRF and proves Eq.~\eqref{eq:mrfcir2}. The result in Eq.~\eqref{eq:mrfcir1} follows analogously.
\end{proof}

\section{FRL-vs-Brier-Score Curves}
\begin{figure}[htp!]
\centering
\includegraphics[width=0.55\textwidth]{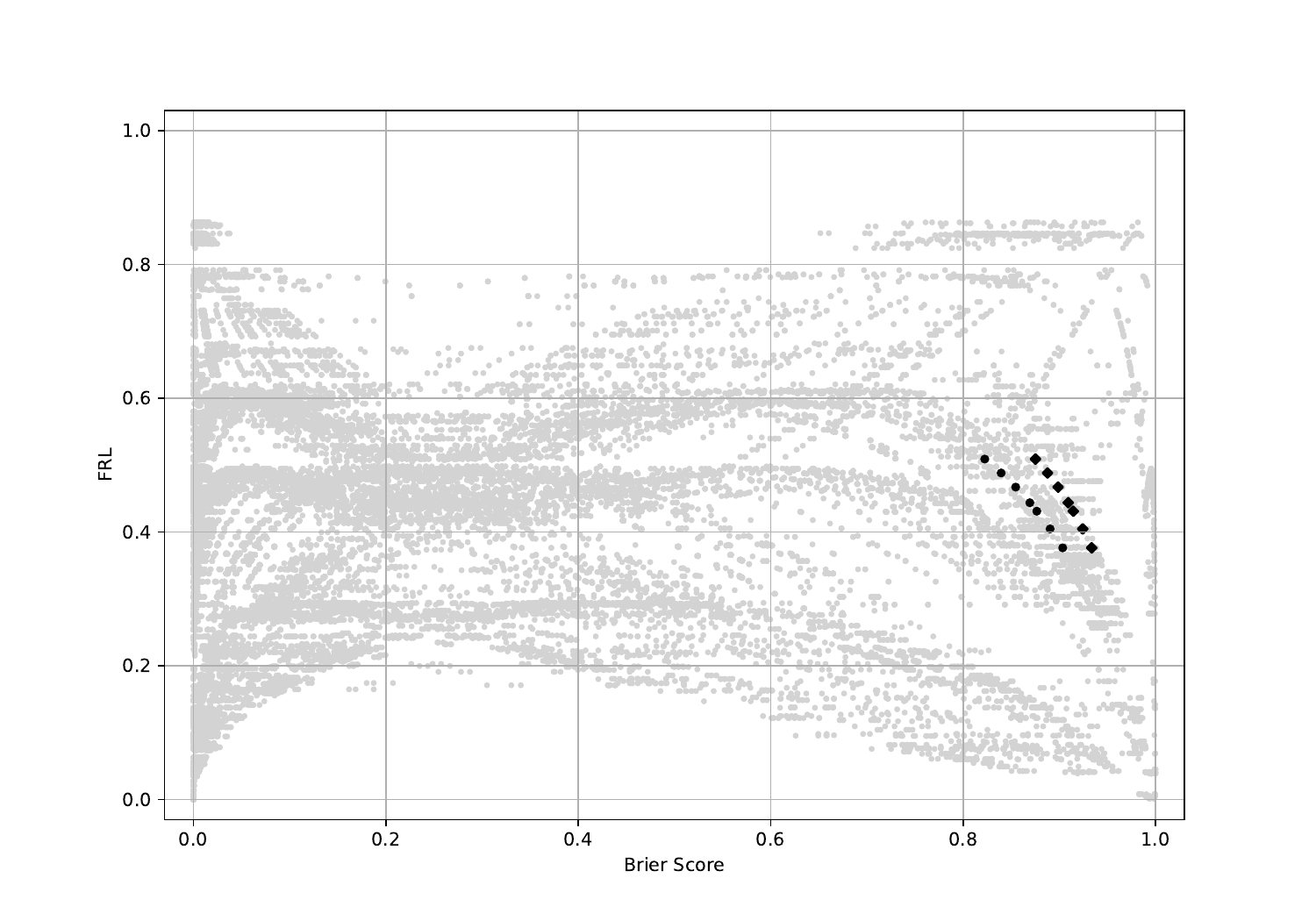}
\caption{A FRL-vs-brier-score scatter plot for the the {\tt Bank marketing} dataset.\label{fig:bank_marketing_emp}}
\end{figure}

The scatter plot in Figure~\ref{fig:bank_marketing_emp} details the output of an experiment with the {\tt Bank market\-ing} dataset in terms of relation between FRL levels and Brier scores. Here, as well as in similar plots we obtained for the other datasets, the points seem to lie on different curves. In the figure, two groups of seven points, each belonging to a different curve, have been highlighted in black. The points in a same group have the same value of the private feature \emph{marital status} and are obtained by changing the public features. Namely, the points with higher Brier scores are those associated with \emph{single}, and the other ones with \emph{married}. We also observe that each point in a group has a corresponding point in the other group with the same FRL value. These pairs correspond to instances with the same public states and different private state. The reason is that, as a straightforward consequence of the definition, two instances with the same public feature have the same FRL level. 

\newpage
\bibliographystyle{amsplain}
\bibliography{biblio}

\end{document}